\documentclass[twoside,leqno]{article}  
\usepackage{SDM} 
\usepackage[flushleft]{threeparttable}
\usepackage{amssymb}
\usepackage{multirow}
\usepackage{mathtools}
\usepackage{wasysym}
\usepackage{float}
\usepackage{morefloats}
\usepackage{comment}
\usepackage{algorithm}
\usepackage[algo2e]{algorithm2e}
\usepackage{algpseudocode}
\usepackage{subfloat}
\usepackage{graphicx}
\usepackage{caption}
\usepackage{subcaption}
\usepackage[english]{babel}
\usepackage{cleveref}
\usepackage{arydshln}
\usepackage{lipsum}
\usepackage{etoolbox}
\usepackage{cleveref}
\usepackage{tikz}
\usepackage{cite} %% for citation
\usetikzlibrary{plotmarks}
\newcommand{\ie}{\emph{i.e.}, }

\crefformat{equation}{(#1)}
\crefformat{figure}{Figure #1}
\crefformat{table}{Table #1}
\crefformat{algorithm}{Algorithm #1}
\usepackage[font=small,labelfont=bf]{caption}
\DeclareCaptionFont{tiny}{\tiny}
\providecommand{\keywords}[1]{\textbf{\textit{Keywords---}} #1}

\begin{document}

\title{\Large Filter based Taxonomy Modification for Improving Hierarchical Classification}
\author{Azad Naik\thanks{Department of Computer Science, George Mason University.
              Email: anaik3@gmu.edu; rangwala@cs.gmu.edu} \\
\and
Huzefa Rangwala\footnotemark[2]}
\date{}

\maketitle

\begin{abstract}
Hierarchical Classification (HC) is a supervised learning problem where unlabeled instances are classified into
a taxonomy of classes. Several methods
that utilize the hierarchical structure have been developed to
improve the HC performance. However, in most cases apriori defined hierarchical structure by domain experts is inconsistent; as a consequence performance improvement is not noticeable in comparison to flat classification methods. We 
propose a scalable data-driven filter based rewiring approach to modify an expert-defined hierarchy. Experimental comparisons of top-down HC with our modified hierarchy, on a 
wide range of datasets shows classification performance 
improvement over the baseline hierarchy ($i.e.$, defined by expert), clustered hierarchy and flattening based hierarchy modification approaches. In comparison to 
existing rewiring approaches, our developed method ($rewHier$) is computationally efficient, enabling it to scale to datasets with large numbers of classes, instances and features. We 
also show that our modified hierarchy leads to improved classification performance
for classes with few training samples in comparison to flat and state-of-the-art HC approaches. 
Source code available for reproducibility at: www.cs.gmu.edu/$\sim$mlbio/TaxMod
\end{abstract}


\section{Introduction}
Taxonomy (hierarchy) is most commonly used to organize large volumes of data.
%It provides a structured view of the domain where classes
%(categories) are organized from the most generic to the most
%specific in a top-down order.
 It has been successfully used in different application domains such as bioinformatics\footnote{http://geneontology.org/}, 
computer vision\footnote{http://www.image-net.org/} and web directories\footnote{http://dir.yahoo.com/}. 
These application domains (especially highlighed
by interest in online prediction challenges such as LSHTC\footnote{http://lshtc.iit.demokritos.gr/} and BioASQ\footnote{http://bioasq.org/})
introduce unique computational and statistical challenges. Given that these datasets have several thousand classes, the developed methods 
need to scale during the learning and prediction phases. Further, the majority of classes have very few training examples, leading to a class imbalance problem where the learned models (for rare categories) have a tendency to overfit and mispredictions favor the larger classes. 

Hierarchies provide useful structural relationships (such as
parent-child and siblings) among different classes that can
be exploited for learning generalized classification models. In the past, researchers have demonstrated the usefulness
of hierarchies for classification and have obtained promising
results \cite{cai2004hierarchical,koller1997hierarchically,mccallum1998improving,dumais2000hierarchical,sun2001hierarchical}. 
Utilizing the hierarchical structure has been shown to improve 
the classification performance 
for rare categories as well \cite{liu2005site}. 
% Why top-down methods
Top-down HC methods that leverage the hierarchy during the
learning and prediction process are effective
approaches to deal with large-scale problems \cite{koller1997hierarchically}. 
Classification decision for top-down methods involves invoking only the models 
in the relevant path within the hierarchy. Though
computationally efficient, 
these methods have higher number of misclassifications due to error 
propagation \cite{naik2016inconsistent}. 

\begin{figure*}[ht!]
    \centering
    \begin{subfigure}[b]{0.445\textwidth}
        \centering
        \includegraphics[width=\textwidth, height=8.5cm]{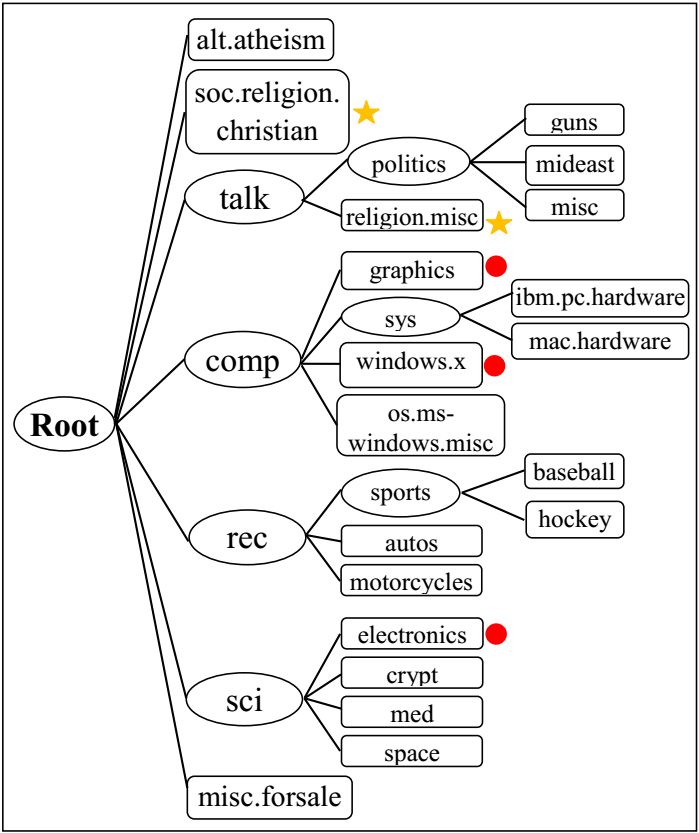}
        \caption{{Expert-defined (Original) Hierarchy}}
    \end{subfigure}
    \hfill
    \begin{subfigure}[b]{0.445\textwidth}
        \centering
        \includegraphics[width=\textwidth, height=8.5cm]{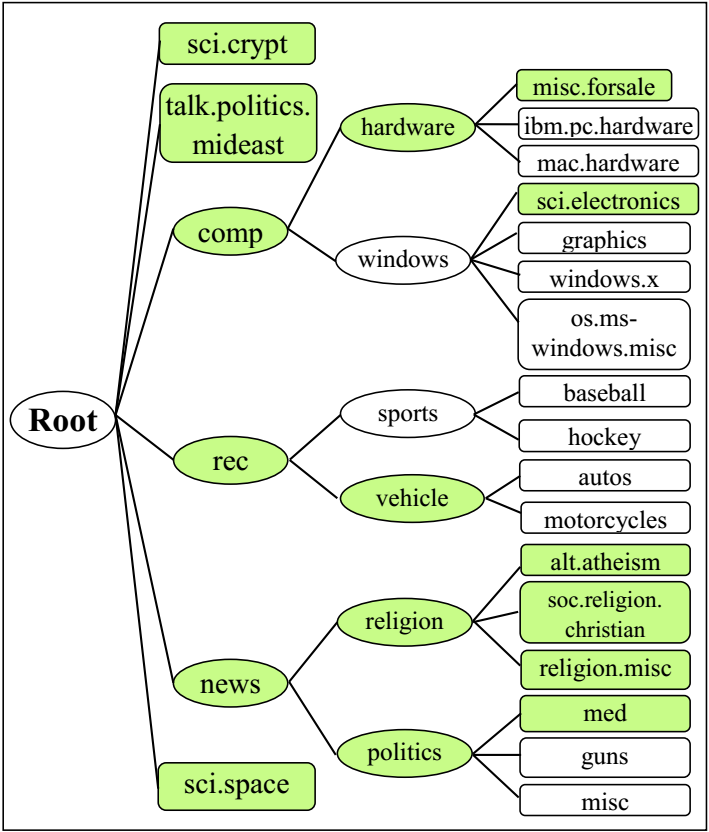}
        \caption{{Clustered Hierarchy}}
    \end{subfigure}
    \hfill
    \begin{subfigure}[b]{0.445\textwidth}
        \centering
        \includegraphics[width=\textwidth, height=8.5cm]{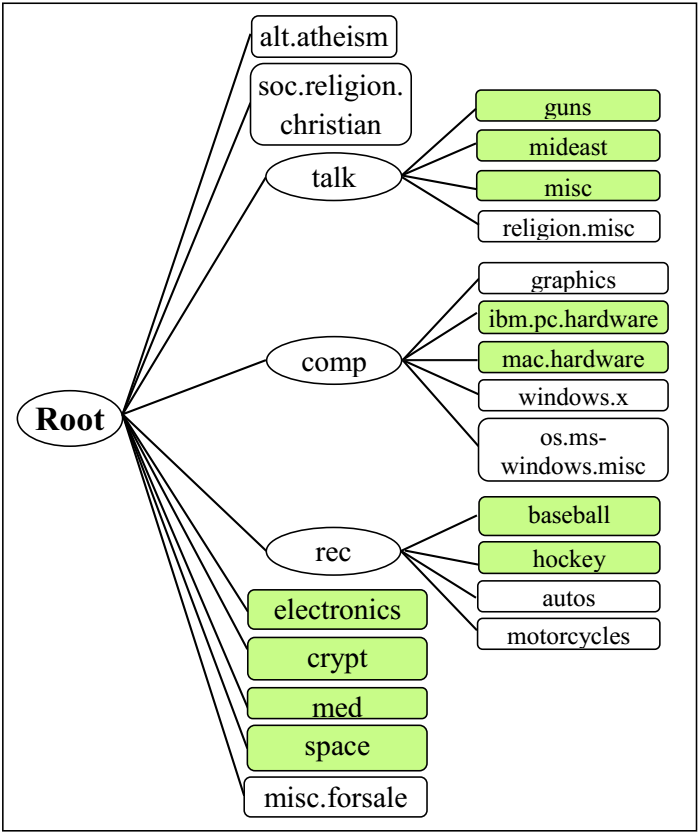}
        \caption{{Flattened Hierarchy}}
    \end{subfigure}
    \hfill
    \begin{subfigure}[b]{0.445\textwidth}
        \centering
        \includegraphics[width=\textwidth, height=8.5cm]{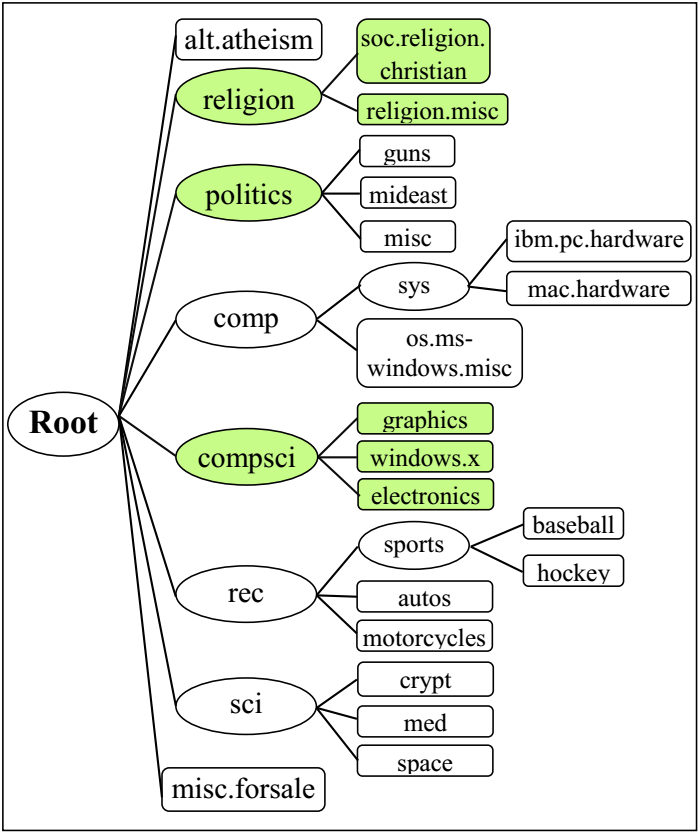}
        \caption{{Rewired Hierarchy}}
    \end{subfigure}
    \caption{{(a) Expert-defined hierarchy (classes with high degree of similarities are marked with symbols 
\begingroup\color{orange}$\bigstar$\endgroup, \begingroup\color{red}$\CIRCLE$\endgroup) modified using various methods: (b) Agglomerative clustering with cluster cohesion to restrict the height to original height \cite{li2007hierarchical} (c) Global-INF flattening method \cite{naik2016inconsistent} (d) Proposed rewiring method. Modified structure changes are shown in green color.}}
    \label{NGHierarchyMod}
\end{figure*}

% Reason for hierarchical inconsistency?
For several  benchmarks, the HC approaches are outperformed by \emph{flat} classifiers 
that ignore the hierarchy \cite{xiao2011hierarchical, zimek2010study}. 
In majority of the cases, the hierarchy available for training classifiers is manually designed by 
experts based on domain knowledge and is not consistent for classification. In order to improve performance, we need 
to \emph{restructure} the hierarchy to make it more favorable and useful for classification. Motivated by this idea, our main focus in this paper is on generating an improved
representation from the expert-defined hierarchy. To summarize, our contributions are as follows:
\begin{itemize}
\vspace*{-2mm}\item We propose an efficient data-driven filter based rewiring approach for hierarchy modification which unlike previous wrapper based
approaches \cite{tang2006acclimatizing,qi2011hierarchy} does not require multiple, expensive computations. Our approach is scalable and
can be applied to the HC problems with high-dimensional features, large number of classes and examples.
\vspace*{-2mm}\item We perform extensive empirical evaluations and case studies to show the strengths of our approach in comparison to other hierarchy modification approaches such as clustering and flattening.
\vspace*{-2mm}\item The modified hierarchy can be used with any hierarchical classification approaches like top-down HC or state-of-the-art approaches incorporating hierarchical relationships \cite{charuvaka2015}. The modified 
hierarchy in conjunction with a scalable Top-Down HC approach outperforms  the flat classifiers on $\sim$65\% of the  rare categories (i.e., classes with less than 10 training examples) across the DMOZ datasets (See Section \ref{flatCost}). 

%We demostrate the effectiveness of our modified hierarchy irrespective of the classifiers being trained such as top-down or
%state-of-the-art cost sensitive learning approaches \cite{charuvaka2015}.   The use of modified hierarchy with HC approaches assists in learning better models for rare categories. Specifically, in comparison to flat approach
%$\sim$65\% of the DMOZ classes  shows improved results 
%with top-down approach and  it increases to $\sim$80\% with cost-sensitive-learning approach when using the modified hierarchy 
%over the expert-defined hierarchy.

\end{itemize}

\begin{table}[t] 
\centering 
\begin{tabular}{| l | c | c | c |} 
\hline
 \multicolumn{1}{|c|}{\bf{Modification Method}} & {\bf{Approach}} & {\bf{Type}} & \multicolumn{1}{c|}{\bf{Scalable}}\\
\hline
{Margin-based modification \cite{babbar2013maximum}} & Flattening & Filter & \multicolumn{1}{c|}{$\checkmark$}\\
{Level flattening \cite{wang2010flatten}} & Flattening & Filter & \multicolumn{1}{c|}{$\checkmark$}\\
{Inconsistent node flattening \cite{naik2016inconsistent}} & Flattening & Filter & \multicolumn{1}{c|}{$\checkmark$}\\
{Learning based algorithm \cite{babbar2013flat}} & Flattening & Wrapper & \multicolumn{1}{c|}{$\times$}\\
{Agglomerative clustering \cite{li2007hierarchical}} & Clustering & Wrapper & \multicolumn{1}{c|}{$\times$}\\
{Divisive clustering \cite{punera2005automatically}} & Clustering & Wrapper & \multicolumn{1}{c|}{$\times$}\\
{Optimal hierarchy search \cite{tang2006acclimatizing}} & Rewiring & Wrapper & \multicolumn{1}{c|}{$\times$}\\
{Genetic based algorithm \cite{qi2011hierarchy}} & Rewiring & Wrapper & \multicolumn{1}{c|}{$\times$}\\
\hline
\bf{{Our proposed approach:}}&&&\\
{Similarity based modification} & Rewiring & Filter & $\checkmark$\\
\hline
\end{tabular}
\caption{{The summary review of existing taxonomy modification methods and their characteristics.}} 
\label{table:litReview} 
\end{table}

\section{Methods} 
\label{defnot}
\subsection{Motivation}
The manual process of hierarchy creation suffers from
various issues. Specifically, (i) Hierarchies are generated by
grouping semantically similar categories under a common
parent category. However, many different semantically sound
hierarchies may exist for same set of classes. For example, in
categorizing products, the experts may generate a hierarchy
by first separating products based on the company name
($e.g.$, Apple, Microsoft) and then the product type ($e.g.$,
phone, tablet) or vice-versa. Both hierarchies are equally good
from the perspective of an expert. However, these different
hierarchies may lead to different classification results. (ii) Apriori it is not clear to domain experts when to generate new
nodes (hierarchy expansion) or merge two or more nodes
(link creation) while creating hierarchies, resulting in a certain
degree of arbitrariness. (iii) A large number of categories pose
a challenge for the manual design of a consistent hierarchy.
(iv) Dynamic changes may require hierarchical restructuring.

To remove inconsistencies, various approaches 
for hierarchy modification have been proposed. These approaches 
  can be broadly categorized into two categories:  (i) Flattening approaches  \cite{naik2016inconsistent,wang2010flatten,babbar2013flat,babbar2013maximum,malik2010improving} where some of the identified inconsistent nodes (based on error rate, classification margins) are flattened (removed) and (ii) Rewiring approaches \cite{tang2006acclimatizing,qi2011hierarchy,nitta2010improving} where parent-child relationships within the  hierarchy are  modified to improve the classification performance. Clustering based methods have also been adapted in some of these studies \cite{li2007hierarchical,punera2005automatically} where consistent hierarchy is generated from scratch using agglomerative or divisive clustering algorithms. A summary of the various existing methods and their characteristics is shown in Table \ref{table:litReview}.

To understand the qualitative difference between hierarchy generated using various approaches, we performed experiments on the smaller newsgroup\footnote{http://qwone.com/$sim$~jason/20Newsgroups/} dataset containing 20 classes. Figure \ref{NGHierarchyMod}(b)-(d) shows the hierarchy structure obtained using clustering, flattening and rewiring based approaches, respectively. Hierarchy generated using clustering completely ignores the expert-defined hierarchy information, which contains valuable prior knowledge for classification \cite{tang2006acclimatizing}. Flattening approaches  cannot
group together the classes from different hierarchical branches (for $e.g$, $soc.religion.christian$ and $religion.misc$). On the contrary, the rewiring approaches provide 
the flexibility of grouping classes from different sub-branches.  More details about Figure \ref{NGHierarchyMod} are discussed later in a case study (Section \ref{caseStudy}).

\begin{table}[t!] 
\centering 
\begin{tabular}{| c| l|} 
\hline
\textbf{Symbol} & \multicolumn{1}{c|}{\textbf{Description}}\\
\hline
$\mathcal{H}$ & expert-defined (original) hierarchy\\
$\mathcal{L}$ & set of leaf categories (classes)\\
${\bf{x}}_i$ & input vector for $i$-th training example\\
$y_i \in \mathcal{L}$ & true label for $i$-th training example\\
$\hat{y}_i \in \mathcal{L}$ & predicted label for $i$-th test example\\
$y_i^n \in \pm 1$ & binary label used for $i$-th training example to learn weight\\
& vector for $n$-th node in the hierarchy. $y_i^n$ = 1 iff $y_i$ =$ n$, -1\\
&  otherwise\\
$\hat{y}_i^n \in \pm 1$ & predicted label for $i$-th test example at $n$-th node in the\\
& hierarchy, $\hat{y}_i^n$ = 1 iff prediction, -1 otherwise\\
$N$ & total number of training examples\\
${\bf{\Theta}}_n$ & weight vector (model) for $n$-th node\\
$f^*_n$ & optimal objective function value for $n$-th node obtained\\
& using validation dataset. We have dropped the subscript $n$ \\
& at some places  for ease of description\\
C $>$ 0 & misclassification penalty parameter\\
$\mathcal{H}_M$ & modified hierarchy obtained after rewiring\\
$\mathcal{N}$ & set of all nodes in the hierarchy (except root)\\
$\tau$ & threshold for grouping similar classes in rewiring method\\
$\pi(n)$ & parent of the n-th node\\
$\mathcal{C}(n)$ & children of the n-th node\\
$\zeta(n)$ & siblings of the n-th node\\
\hline
\end{tabular}
\caption{{Notation.}} 
\label{table:notDes} 
\end{table}

\subsection{Proposed Rewiring Approach}
\label{proMethod}
\begin{algorithm}[t!]
\SetAlgoLined
 \KwData{Original Hierarchy $\mathcal{H}$, input-output (${\bf{x}}_i, y_i$)}
 \KwResult{Modified Hierarchy $\mathcal{H}_M$}

   /* \textbf{Initialization} */

   $H_M$ = $\mathcal{H}$;

   /* \textbf{Ist step: Grouping Similar Classes Pair} */

   Compute cosine similarity between all possible class pairs.

   /* \textbf{similar class grouping} */

Identify the most similar class pairs with similarity scores value greater than empirically defined threshold parameter $\tau$. Let $|c|$ denotes the number of such pairs represented by the set $\textbf{S} = \{s_1, s_2, \ldots, s_{|c|}\}$, where $i$-th pair $s_i$ is represented using ($s_i^{(1)}, s_i^{(2)}$).

   /* \textbf{IInd step: Inconsistency Identification and Correction} */

   \For{$i$ $=$ 1 to $|c|$}{
   	$rewire[1]=1$;   /* \textbf{check if rewiring is needed for $s_i^{(1)}$} */\\ 
           $rewire[2]=1$; /* \textbf{check if rewiring is needed for $s_i^{(2)}$} */\\ 
           /* \textbf{Inconsistent pair check} */\\
	\If{$\pi(s_i^{(1)}) \neq \pi(s_i^{(2)})$}{  
                      /* \textbf{check similarity to all siblings} */\\
		\ForEach{$j$ $\in$ $\zeta(s_i^{(1)})$}{
			\If{$\big((j, s_i^{(2)})$ or $(s_i^{(2)}, j)\big)$ $\notin {\textbf{S}}$}{
				$rewire[2]=0$;\\
				break;
			}
		}
		\ForEach{$j$ $\in$ $\zeta(s_i^{(2)})$}{
			\If{$\big((j, s_i^{(1)})$ or $(s_i^{(1)}, j)\big)$ $\notin {\textbf{S}}$}{
				$rewire[1]=0$;\\
				break;
			}
		}
		\eIf{$(rewire[1]==0)$ and $(rewire[2]==0)$}{
			/* \textbf{perform node creation} */\\			
			$N_{new}$ = $\phi$ /* \textbf{create new node} */\\
			$[\mathcal{H}_M] = {\bf{NC}}(N_{new} {\rightarrow} lca(s_i), s_i {\rightarrow} N_{new}, \mathcal{H}_M)$;\\			
                                /* \textbf{lca denotes lowest common ancestor} */\\
		}{
			\eIf{$(rewire[1]==1)$}{
  		 		[$\mathcal{H}_M$] = {\bf{PCRewire}}$(s_i^{(1)} {\rightarrow} \pi(s_i^{(2)}), \mathcal{H}_M)$;
  			 }{
   				[$\mathcal{H}_M$] = {\bf{PCRewire}}$(s_i^{(2)} {\rightarrow} \pi(s_i^{(1)}), \mathcal{H}_M)$;
  			}
		}
	}
    }
    /* \textbf{perform node deletion} */

   [$\mathcal{H}_M$]  = {\bf{ND}}$(\mathcal{H}_M)$;

\Return{$\mathcal{H}_M$}
 \caption{\emph{rewHier} Algorithm}
 \label{TAXMOD}
\end{algorithm}

\begin{comment}
Inconsistencies in the hierarchy can deteriorate the performance of the learned models. To improve classification performance, we need to remove these inconsistencies before using hierarchical structure in the learning framework. 
\end{comment}

Wrapper based approaches \cite{tang2006acclimatizing,qi2011hierarchy,nitta2010improving} modify the hierarchy
by making one or few changes, which are then evaluated for
classification performance improvement using the HC learning algorithm. Modified changes are retained if the performance
results improve; otherwise the changes are discarded and the
process is repeated. This repeated procedure of hierarchy
modification continues until the optimal hierarchy that satisfies
certain criteria is reached. As such, wrapper approaches are not scalable for large datasets. 

We propose an efficient data-driven filter based rewiring approach
where the hierarchy is modified based on certain 
relevance criterion (pairwise sibling similarity) between the different classes within the hierarchy. Our approach is single
step and do not require experimental evaluation for 
multiple iterations. We refer to our proposed rewiring approach as $rewHier$. Table \ref{table:notDes} captures the common notations used in this paper and 
Algorithm \ref{TAXMOD} illustrates our approach for hierarchy modification. Specifically, it consists of two  steps:

{\bf{(i) Grouping Similar Classes Pairs}} -  To ensure classes with high degree of similarity are grouped together under the same parent node in the modified taxonomy, this step identifies the similar classes pairs that exist within the expert-defined  hierarchy. Pairwise cosine similarity is used as the similarity measure in our experiments because it is less prone to the curse of dimensionality \cite{steinbach2004challenges}.  Once the similarity scores are computed, we determine the set $\textit{\bf{S}}$ of most similar pairs of classes using an empirically defined cut-off threshold $\tau$ for a dataset (detailed analysis regarding $\tau$ selection is discussed in Section \ref{empDef}). For example, in Figure \ref{NGHierarchyMod}(a) this step will group together the class pairs with high similarity scores such as $\textit{\bf{S}}$ = \big[($religion.misc$, $soc.religion.christian$), ($electronics$, $windows.x$), ($electronics$, $graphics$), $\cdot$$\cdot$$\cdot$\big].

Pairwise similarity computation between different classes is one of the major bottlenecks of this step. To make it scalable, we distribute the similarity computation across multiple compute nodes.

\begin{figure}[t!]
    \centering
    \begin{subfigure}[b]{0.24\textwidth}
        \centering
        \includegraphics[width=\textwidth, height=2.5cm]{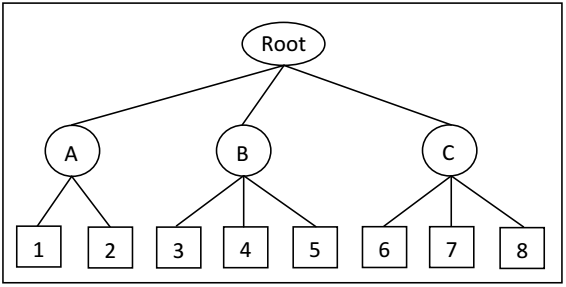}
        \caption{{Expert-defined Hierarchy}}
    \end{subfigure}
    \hfill%
    \begin{subfigure}[b]{0.23\textwidth}
        \centering
        \includegraphics[width=\textwidth, height=2.5cm]{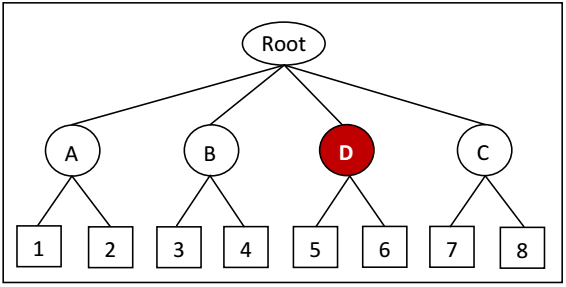}
        \caption{{Node Creation} ({\bf{D}})}
    \end{subfigure}
    \begin{subfigure}[b]{0.24\textwidth}
        \centering
        \includegraphics[width=\textwidth, height=2.5cm]{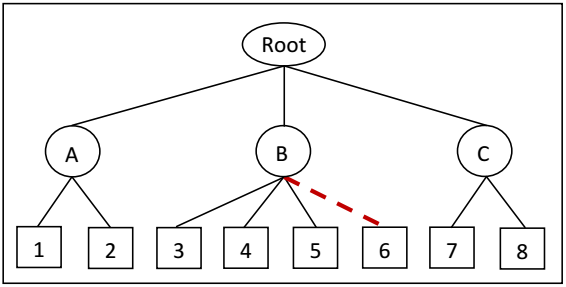}
        \caption{{Parent-child Rewiring} ({\bf{6}})}
    \end{subfigure}
    \hfill%
    \begin{subfigure}[b]{0.23\textwidth}
        \centering
        \includegraphics[width=\textwidth, height=2.5cm]{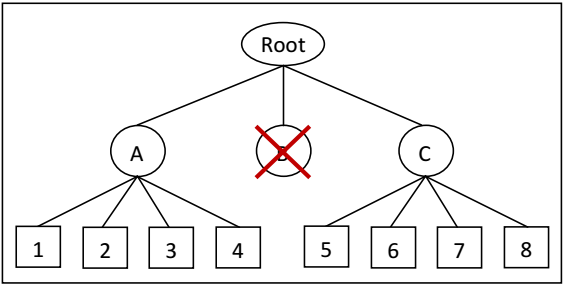}
        \caption{{Node Deletion} ({\bf{B}})}
    \end{subfigure}
    \caption{{{Modified hierarchical structures (b)-(d) obtained after applying elementary operations to expert-defined hierarchy ($\mathcal{H}$). Leaf nodes are marked with `rectangle' and structural changes are shown by red color.}}}
    \label{incLink}
\end{figure}

{\bf{(ii) Inconsistency Identification and Correction}} - To obtain the consistent hierarchy, we group together each of the similar class pairs to a common parent node. Iteratively, starting from the most similar class pairs we check for potential inconsistencies  $i.e.$, if the pairs of classes are in different branches (sub-trees). In order to resolve the identified inconsistencies we take  corrective measures using three basic elementary operations: (i) node creation, (ii) parent-child rewiring and (iii) node deletion. Figure \ref{incLink}(b)-(d) illustrates the various hierarchical structures that are obtained after the execution of these elementary operations on the 
expert-defined hierarchy in  Figure \ref{incLink} (a).\\
 \hspace*{5mm}{\textbf{Node Creation (NC)}} - This operation groups together the identified similar class pairs in different branches (sub-trees) of the hierarchy using a new node, with parent as the lowest common ancestors of similar classes.  Figure \ref{incLink}(b) illustrates this operation where the similar class pairs $\bf{5}$ and $\bf{6}$  are grouped together by the newly created node $\bf{D}$. This operation is used only when a  \emph{proper subset} of the leaf nodes from different branches are similar ($i.e.$, not similar to all leaf nodes in the branch; otherwise the parent-child rewiring operation is used).\\
 \hspace*{5mm}{\textbf{Parent-child Rewiring (PCRewire)}} - As shown in Figure \ref{incLink}(c), this operation simply assigns (rewires) the leaf node from one parent to another parent node in the hierarchy. It is useful when the leaf node is identified to be similar to all the sibling leaf nodes within the given hierarchy branch. For example, in Figure \ref{incLink}(c), if the computed similarity score determines the leaf node $\bf{6}$ to be more similar to nodes $\bf{3}$, $\bf{4}$ and $\bf{5}$ in comparison to its current siblings $\bf{7}$ and $\bf{8}$, than it is more desirable from a classification perspective to assign $\bf{6}$ as node ${\bf{B}}$ child rather than ${\bf{C}}$.\\
 \hspace*{5mm}{\textbf{Node Deletion (ND)}} - This refers to deletion of nodes in the hierarchy that are deemed useless for classification. In Figure \ref{incLink}(d), node $\bf{B}$ is deleted because  there are no leaf nodes that can be classified by node $\bf{B}$. This operation is used as a post-processing step in our algorithm to refine the hierarchy.

The $rewHier$ algorithm determines (outer \emph{for loop}) the best corrective
measures (\emph{node creation} or \emph{parent-child rewiring}) that need to be taken. Once all the inconsistencies 
have been addressed, $rewHier$ calls the \emph{node deletion} procedure as a final modification step where unnecessary nodes are deleted. 

It should be noted that the  new modified hierarchy obtained after inconsistencies
removal can be used to train any HC classifier.
State-of-the-art HC classification approaches embed the parent-child relationships from the hierarchy either, 
within the regularization term \cite{gopal2013recursive},
referred by HR-LR (or HR-SVM) or the 
loss term, referred by HierCost \cite{charuvaka2015}.
The intuition behind Hierarchy Regularized Logistic Regression (HR-LR) \cite{gopal2013recursive} approach 
is that data-sparse child nodes benefit  during training 
from data-rich parent nodes, and this has been shown 
to achieve the best performance on standard HC benchmarks. However, training 
these models is computationally expensive  due 
to the coupling between different classes within this  formulation. 
To make this method scalable, 
distributed computation using hadoop map-reduce was proposed in 
conjunction with parallel training of odd and even levels. As such,
this method requires special hardware and software configurations for 
large datasets and hence, we did not use this method in our experiments. 
In case of HierCost \cite{charuvaka2015}, a cost-sensitive learning approach 
was adapted. This method intuitively captures the hierarchical information 
by treating misclassifications differently based on the commonalities (ancestors) 
between 
the true and the predicted labels. Intrinsically, this method scales for 
large datasets due to the trivial decomposition of learned models for different leaf 
categories. This method  outperforms 
HR-LR 
method without any additional parameter 
configurations. Hence, in this paper we use the HierCost 
approach for evaluation with our rewired hierarchies. 

\subsection{Top-Down Hierarchical Classification}
We propose to use the Top-Down HC approach  with our modified hierarchies because it scales well during training and prediction. 
Specifically, we train binary one-vs-rest classifiers for each of the nodes $n\in\mathcal{N}$ --- to discriminate its positive examples from the
examples of other nodes (\ie negative examples) in the hierarchy. In this paper, we use logistic regression (LR) as the underlying
base model for training \cite{gopal2013recursive}.
The LR objective uses logistic loss to minimize the empirical risk and squared
$l2$-norm term (denoted by $||\cdot||_{2}^{2}$) to control model
complexity and prevent overfitting. The objective function $f_n$ for training
a model corresponding to node $n$ is provided in eq. (2.1).
\begin{equation}
f_n=\min_{{\bf{\Theta_{n}}}}\Bigg[C\sum_{i=1}^{N}\log\left(1+\exp\left(-y_{i}^{n}{\bf{\Theta_{n}}}^{T}\mathbf{x}_i\right)\right)+\frac{1}{2}\left\Vert {\bf{\Theta_{n}}}\right\Vert _{2}^{2}\Bigg]\label{ARLR}
\end{equation}

For each node $n$ in the hierarchy, we solve eq. (2.1) to obtain the optimal weight
vector denoted by ${\bf{\Theta}}_{n}$. The complete set of parameters for
all the nodes $[{\bf{\Theta}}_{n}] _{n\in\mathcal{N}}$ constitutes
the learned model for top-down classifier. For LR
models, the conditional probability for $\hat{y}_i^{n}\in\pm1$ given
its feature vector $\textbf{x}_i$ and the weight vector ${\bf {\Theta}}_n$
is given by eq. (2.2) and the decision function by eq. (2.3).
\begin{align}
P\big(\hat{y}_i^{n}\mid{{\bf{x}}_i},{\bf{\Theta}}_n\big)=\begin{alignedat}{1} &  \mathlarger{1}\big/{\left(1+\exp\left(-y_{i}^{n}{\bf{\Theta}}_{n}^{T}{\bf{x}}_i\right)\right)}
\end{alignedat}
\label{APD}
\end{align}
\begin{align}
\hat{y}_i^{n}=\left\{ \begin{alignedat}{1} & +1 \quad f_n({\bf{x}}_i) = {\bf{\Theta}}_{n}^{T}{\bf{x}}_i\ge0\\
 & -1 \quad \text{otherwise}
\end{alignedat}
\right\} \label{eq:ADEC}
\end{align}

For a test example with feature vector ${\bf{x}}_i$, the top-down
classifier predicts the class label $\hat{y}_i\in\mathcal{L}$ as shown
in eq. (2.4). Essentially, the algorithm starts at
the root and recursively selects the best child node until it
reaches a terminal node which is the predicted label.
\begin{align}
\hat{y}_i=\left\{ \begin{alignedat}{1} & \mathbf{initialize}\quad p:=root\\
 & \mathbf{while}\ p\notin\mathcal{L}\\
 & \quad p:=\mathbf{argmax}_{q\in\mathcal{C}(p)}\ f_{q}({\bf{x}}_i)\\
 & \mathbf{return}\ p
\end{alignedat}
\right\} \label{eq:TDTest}
\end{align}

\section{Experimental Protocol}
\label{expResults}
\subsection{Datasets}
We have used an extensive set of datasets for 
evaluating the performance of our proposed rewiring approach. Various statistics of the datasets 
used are listed in Table \ref{table:finaltabledataset}. CLEF \cite{dimitrovski2011hierarchical} and DIATOMS \cite{Dimitrovski11:jrnl} are image datasets and the rest are 
text datasets. IPC\footnote{http://www.wipo.int/classifications/ipc/en/}  is a collection of patent documents and the DMOZ datasets are
an archive of web-pages available from LSHTC\footnote{http://lshtc.iit.demokritos.gr/} challenge website. For evaluating the DMOZ-2010 and DMOZ-2012 datasets we  use
the provided test split. The results reported for these two benchmarks are blind prediction (i.e., we do not know the ground truth labels for the test set) 
obtained from the 
web-portal interface\footnote{http://lshtc.iit.demokritos.gr/node/81}$^{,}$\footnote{ http://lshtc.iit.demokritos.gr/LSHTC3\_oracleUpload}. For all text datasets we apply
the tf-idf transformation with $l2$-norm normalization on the word-frequency feature vector. 

\begin{table}[t!] 
\centering 
\begin{tabular}{| l | c c c c c c |} 
\hline
 \multicolumn{1}{|c|}{\multirow{2}{*}{\bf{Dataset}}} & {\bf{Total}} & \multicolumn{1}{c}{\bf{Leaf}} & \multicolumn{1}{@{}c@{}}{\multirow{2}{*}{\bf{Levels}}} & \multicolumn{1}{c}{\multirow{2}{*}{\bf{Train}}} & \multicolumn{1}{c}{\multirow{2}{*}{\bf{Test}}} & \multicolumn{1}{c@{}|}{\multirow{2}{*}{\bf{Features}}}\\
\multicolumn{1}{|c|}{} & {\bf{Nodes}} & {\bf{Nodes}} &  &  &  & \\
\hline
\bf{CLEF} & 88 & 63 & 3 & 10000 & 1006 & 80\\
\bf{DIATOMS} & 399 & 311 & 3 & 1940 & 993 & 371\\
\bf{IPC} & 553 & 451 & 3 & 46324 & 28926 & 1123497\\
\bf{DMOZ-SMALL} & 2388 & 1139 & 5 & 6323 & 1858 & 51033 \\
\bf{DMOZ-2010} & 17222 & 12294 & 5 & 128710 & 34880 & 381580\\
\bf{DMOZ-2012} &  13963 & 11947 & 5 & 383408 & 103435 & 348548\\
\hline
\end{tabular}
\caption{{Dataset statistics.}} 
\label{table:finaltabledataset} 
\end{table}

\subsection{Evaluation Metrics}~\\
  \hspace*{5mm}{\textbf{Flat Measures}} - Micro-$F_1$ ($\mu$$F_1$) and Macro-$F_1$ (M$F_1$) are 
  used for evaluating the performance. To compute $\mu$$F_1$, we sum up the category specific true positives $(TP_c)$, false positives $(FP_c)$ and false negatives $(FN_c)$ for different classes and compute the score as:
\begin{gather}
P = \frac{\sum_{c \in \mathcal{L}}TP_c}{\sum_{c \in \mathcal{L}}(TP_c + FP_c)}, R = \frac{\sum_{c \in \mathcal{L}}TP_c}{\sum_{c \in\mathcal{L}}(TP_c + FN_c)} \nonumber \\
\mu F_1 = \frac{2PR}{P + R}\nonumber\end{gather}

Unlike $\mu$$F_1$ that gives equal weight to each instance, M$F_1$ gives equal weight to all classes so that score is not skewed in favor of the larger classes. It is computed as: 
\begin{gather} 
P_c = \frac{TP_c}{TP_c + FP_c}, R_c = \frac{TP_c}{TP_c + FN_c} \nonumber \\
MF_1 = \frac{1}{|\mathcal{L}|}\sum_{c \in \mathcal{L}}\frac{2P_cR_c}{P_c + R_c}\nonumber 
\end{gather}
\\
 \hspace*{5mm}{\textbf{Hierarchical Measures}} - Hierarchy is used for evaluating the classifier performance. The hierarchy based measure include hierarchical $F_1$ ($hF_1$) defined by:
\begin{gather}
hP = \frac{\sum_{i=1}^{N}|{\mathcal{{A}}}(\hat{y_i}) \cap {\mathcal{{A}}}({y_i})|}{\sum_{i=1}^{N}|{\mathcal{{A}}}(\hat{y_i})|}, hR = \frac{\sum_{i=1}^{N}|{\mathcal{{A}}}(\hat{y_i}) \cap {\mathcal{{A}}}({y_i})|}{\sum_{i=1}^{N}|{\mathcal{{A}}}({y_i})|} \nonumber\\
h F_1 = \frac{2*hP*hR}{hP + hR}\nonumber\end{gather}
where ${\mathcal{{A}}}(\hat{y_i})$ and ${\mathcal{{A}}}({y_i})$ are  the sets of ancestors of the predicted and true labels which include the label itself, but do not include the root node, respectively.

Note that for consistent evaluation, we have used the original hierarchy for all methods unless noted.

\subsection{Methods for Comparison}
\subsubsection{Hierarchical Methods} 
\label{TDbaselinemethods}
Based on the hierarchy used during the training process, we use the following methods for comparison.\\
 \hspace*{5mm}{\textbf{Top-Down Logistic Regression (TD-LR)}}: Expert-defined hierarchy provided by domain experts is used for training the classifiers.\\
  \hspace*{5mm}{\textbf{Clustering Approach}}: Hierarchy generated using agglomerative clustering is used. For evaluation, we have restricted the height of clustered hierarchy to the original height by flattening using cluster cohesion \cite{li2007hierarchical}.\\
  \hspace*{5mm}{\textbf{Global Inconsistent Node Flattening (Global-INF)}} \cite{naik2016inconsistent}: Hierarchy is modified by flattening (removing) the  inconsistent nodes based on optimal optimization objective value obtained at each node (eq. (2.1)) and empirically defined global cut-off threshold.\\
  \hspace*{5mm}{\textbf{Optimal Hierarchy Search}} \cite{tang2006acclimatizing}: Optimal hierarchy is
identified in the hierarchical space by gradually modifying the expert-defined hierarchy using elementary operations -- promote, demote and merge. For reducing the number of operations (and hence hierarchy evaluations), we have restricted the modification to the hierarchy branches where we encountered the maximum classification errors. This modified approach is referred as T-Easy. In the original paper \cite{tang2006acclimatizing}, the
largest evaluated dataset has 244 classes and 15795 instances.
\subsubsection{Flat Method} The hierarchy is ignored and  binary one-versus-rest $l2$-regularized LR classifiers are trained 
for each of the leaf categories. The prediction decision for unlabeled test instances is based on the maximum prediction score achieved  across the  several  leaf categories classifiers.
\subsubsection{State-of-the-art Cost-sensitive Learning} \cite{charuvaka2015} 
Similar to flat method but with cost value associated with each instance in the loss function as shown in eq. (3.5). This approach is referred as HierCost and for evaluations we have used the best cost function ``exponential tree distance (ExTrD)'' proposed in the paper.
\begin{equation}
f_n=\min_{{\bf{\Theta_{n}}}}\Bigg[C\sum_{i=1}^{N}\sigma_i\log\left(1+\exp\left(-y_{i}^{n}{\bf{\Theta_{n}}}^{T}\mathbf{x}_i\right)\right)+\frac{1}{2}\left\Vert {\bf{\Theta_{n}}}\right\Vert _{2}^{2}\Bigg]\label{costSen}
\end{equation}
where $\sigma_i$ is the cost value assigned to example $i$.

\subsection{Experimental Settings}
\label{exp:setting}
To make the experimental results comparable to previously published results we use  the same train-test split as provided by the public 
benchmarks. In all the experiments we divide the training dataset into train and a small validation dataset in the ratio 90:10. The final reported testing performance
is done on an independent held-out dataset as provided by these benchmarks. The model is trained by choosing the misclassification 
penalty parameter $C$ in the set \big[0.001, 0.01, 0.1, 1, 10, 100, 1000\big]. The best parameter 
selected using a validation set is used to retrain the models on the entire training set. For our proposed rewiring approach, we compute the pairwise similarities between classes using the entire training dataset. Additionally, we use the liblinear 
solver\footnote{http://www.csie.ntu.edu.tw/$\sim$cjlin/liblinear/} for optimization in all the experiments. The source code is 
made available at our website: http://cs.gmu.edu/$\sim$mlbio/TaxMod

\begin{table}[t!] 
\centering 
\begin{tabular}{|c|c| c| c| c|} 
\hline
{\bf{}}& \multirow{2}{*}{{\bf{TD-LR}}}& {\bf{Clustering \cite{li2007hierarchical}}} & {\bf{Flattening \cite{naik2016inconsistent}}} & {\bf{Proposed}}\\
{\bf{Metric}}&& {\bf{Agglomerative}} & {\bf{Global-INF }} & {\bf{rewHier}}\\
                    & [Figure \ref{NGHierarchyMod}(a)] & [Figure \ref{NGHierarchyMod}(b)] & [Figure \ref{NGHierarchyMod}(c)] & [Figure \ref{NGHierarchyMod}(d)]\\
\hline
\textbf{$\mu$$F_1 (\uparrow)$} & 77.04 (0.18) & 78.00 (0.09) & 79.42 (0.12) & {\bf{81.24 (0.08)}}\\
\textbf{$MF_1 (\uparrow)$} & 77.94 (0.04) & 78.20 (0.01) & 79.82 (0.07) & {\bf{81.94 (0.04)}}\\
\hline
\end{tabular}
\caption{$\mu$$F_1$ and $MF_1$ performance comparison using different hierarchy modification approaches on newsgroup dataset. Table shows mean and (standard deviation) in bracket across five runs. } 
\label{table:caseStudy} 
\end{table}

\begin{table*}[t!] 
\centering 
\scriptsize
\begin{tabular}{|l |c |c|c|c|c: c|} 
\hline
\multicolumn{1}{|c|}{\multirow{2}{*}{\bf{Dataset}}}& {{\bf{Evaluation}}} & \multicolumn{1}{c|}{\multirow{2}{*}{\bf{TD-LR}}} & \multicolumn{1}{c|}{{\bf{Agglomerative}}} & \multicolumn{1}{c|}{{\bf{Flattening}}} & \multicolumn{2}{c|}{{\bf{Rewiring Methods}}}\\
{} & \bf{Metrics} & & \bf{Clustering \cite{li2007hierarchical}} & \bf{Global-INF \cite{naik2016inconsistent}} & \bf{T-Easy \cite{tang2006acclimatizing}} & \bf{rewHier}\\
\hline
\multirow{2}{*}{\textbf{CLEF}} & \textbf{$\mu$$F_1 (\uparrow)$} & 72.74 & 73.24 & 77.14 & {\bf{78.12}} & 78.00 \\
&\textbf{$M$$F_1 (\uparrow)$} & 35.92 & 38.27 & 46.54 & {\bf{48.83$\blacktriangle$}} & 47.10$\blacktriangle$\\
\multirow{2}{*}{\textbf{DIATOMS}} & \textbf{$\mu$$F_1 (\uparrow)$} & 53.27& 56.08 & 61.31 & {\bf{62.34$\blacktriangle$}} & 62.05$\blacktriangle$ \\
&\textbf{$M$$F_1 (\uparrow)$} & 44.46 & 44.78 & 51.85 & {\bf{53.81$\blacktriangle$}} & 52.14$\blacktriangle$\\
\multirow{2}{*}{\textbf{IPC}} & \textbf{$\mu$$F_1 (\uparrow)$} & 49.32 & 49.83 & 52.30 & 53.94$\vartriangle$ & {\bf{54.28$\vartriangle$  }}\\
&\textbf{$M$$F_1 (\uparrow)$} & 42.51 & 44.50 & 45.65 & {\bf{46.10$\vartriangle$}} & 46.04$\vartriangle$\\
\multirow{2}{*}{\textbf{DMOZ-SMALL}} & \textbf{$\mu$$F_1 (\uparrow)$} & 45.10 & 45.94 & 46.61 & NS & \bf{48.25$\vartriangle$}\\
&\textbf{$M$$F_1 (\uparrow)$} & 30.65 & 30.75 & 31.86 & NS & \bf{32.92$\blacktriangle$}\\
\multirow{2}{*}{\textbf{DMOZ-2010}} & \textbf{$\mu$$F_1 (\uparrow)$} & 40.22 & NS & 42.37&  NS & {\bf{43.10}}\\
&\textbf{$M$$F_1 (\uparrow)$} & 28.37 & NS & 30.41 & NS & {\bf{31.21}}\\
\multirow{2}{*}{\textbf{DMOZ-2012}} & \textbf{$\mu$$F_1 (\uparrow)$} & 50.13 & NS & 50.64  & NS & {\bf{51.82}}\\
&\textbf{$M$$F_1 (\uparrow)$} & 29.89& NS & 30.58  & NS & {\bf{31.24}}\\
\hline
\end{tabular}
\caption{{$\mu$$ F_1$ and $M$$F_1$ performance comparison using different hierarchy modification approaches. $\blacktriangle$ ($\vartriangle$) indicates that improvements are statistically significant with 0.01 (0.05) significance level. We have used sign-test and non- parameteric wilcoxon rank test for statistical evaluation of $\mu F_1$ and $MF_1$ scores, respectively. Test are performed between rewiring methods and the best baseline, Global-INF. These statistical tests are not performed on DMOZ-2010 and DMOZ-2012 datasets because we do not have access to true labels from the online evaluation system. `NS' denotes Not Scalable.}} 
\label{table:HPC} 
\end{table*}

\section{Discussion and Results}
\subsection{Case Study}
\label{caseStudy}
To understand the quality of different hierarchical structures (expert-defined, clustered, flattened and rewired) for the newsgroup dataset shown in Figure \ref{NGHierarchyMod}, we perform top-down HC using 
each of the hierarchy, separately. The dataset has 11269 training instances, 7505 test instances and 20 classes. We evaluate each of the hierarchy 
by randomly selecting five different sets of training and test split in the same ratio as original dataset. %For consistent evaluation across different hierarchies, we chose same set of train/test split. 

The results of classification performance  is shown in Table \ref{table:caseStudy}. We can see  that using these 
modified hierarchies substantially improves the classification performance in comparison to the baseline expert-defined hierarchy. On comparing the clustered, flattened and proposed rewired hierarchies, 
the classification performance obtained from using the rewired hierarchy is found to be significantly better than 
the flattened and clustered hierarchy. This is because rewired hierarchy can resolve inconsistencies by grouping together the classes from different 
hierarchical branches.

\subsection{Evaluating Rewiring Approaches}

\subsubsection{Performance based on Flat Metrics} Table \ref{table:HPC} shows the $\mu F_1$ and $MF_1$ performance comparison of rewiring approaches against expert-defined, clustered and flattened hierarchy baselines. The 
rewiring approaches consistently outperform other baselines for all the datasets across all metrics. For image datasets, the relative performance improvement is larger with performance improvement up to $\sim$11$\%$ using  M$F_1$ scores in  comparison to the baseline TD-LR method. 

In Table \ref{table:HPC} results with $p$-values $<  0.01$ and $<  0.05$ 
are denoted by $\blacktriangle$ and $\vartriangle$, respectively. 
We compute the sign-test for  $\mu F_1$ \cite{yang1999re} and non-parametric wilcoxon rank test
for $MF_1$ comparing the $F_1$ scores obtained per class for the rewiring methods against the best baseline \ie Global-INF. Both, the rewiring approaches significantly outperform the Global-INF method across the different datasets.

The proposed $rewHier$ approach shows competitive classification performance in comparison to the T-Easy approach.
For  smaller datasets, the T-Easy approach has better performance because it searches for the optimal hierarchy in the hierarchical space. However, the main drawback 
of the T-Easy approach is that it requires computationally 
expensive learning-based evaluations for reaching the optimal hierarchy making it  intractable for large, real-world classification benchmarks  such as DMOZ (See detailed discussion in Runtime Comparison). 

\begin{table}[t!] 
\centering
\begin{tabular}{| l | c | c| c:c|} 
\hline
\multicolumn{1}{|c|}{\multirow{2}{*}{\bf{Dataset}}}& {\bf{Hierarchy}} & {\bf{Flattening}}& \multicolumn{2}{c|}{\bf{Rewiring Methods}}\\
{} & \bf{used} & \bf{Global-INF} & \bf{T-Easy \cite{tang2006acclimatizing}} & \bf{rewHier}\\
\hline
\multirow{2}{*}{\textbf{CLEF}} & Original & 79.06 & 81.43  & {80.14} \\
& Modified & 80.87 & 81.82  & 81.28 \\
\multirow{2}{*}{\textbf{DIATOMS}} & Original & 62.80 & 64.28 & 63.24 \\
& Modified & 63.88 & 66.35 & 64.27 \\
\multirow{2}{*}{\textbf{IPC}} & Original & 64.73 & 67.23 & 68.34 \\
& Modified & 66.29  & 68.10 & 68.36 \\ 
\multirow{2}{*}{\textbf{DMOZ-SMALL}} & Original & 63.37 & NS & 66.18 \\
& Modified & 64.97 & NS & 66.30 \\ 
\multirow{1}{*}{\textbf{DMOZ-2012}} & Original & 73.19 & NS & 74.21 \\
\hline
\end{tabular}
\label{table:HPCte} 
\caption{{$hF_1$ performance comparison over expert-defined and new modified hierarchy. For DMOZ-2010 dataset $hF_1$ score is not available from the online evaluation system and for DMOZ-2012 dataset modified hierarchy is not supported.}}
\label{HM}
\end{table}

\begin{table}[t!] 
\centering 
\begin{tabular}{|l|c|c| c:c|} 
\hline
\multicolumn{1}{|c|}{\multirow{2}{*}{\bf{Dataset}}} & {\bf{Baseline}}& {\bf{Flattening}} & \multicolumn{2}{c|}{\bf{Rewiring Methods}}\\
{} & {\bf{TD-LR}}  & {\bf{Global-INF}} & \bf{T-Easy \cite{tang2006acclimatizing}} & \bf{rewHier}\\
\hline
\multirow{1}{*}{\textbf{CLEF}} & 2.5 & 3.5 &  59 & 7.5\\
\multirow{1}{*}{\textbf{DIATOMS}} & 8.5 & 10 & 268 & 24\\
\multirow{1}{*}{\textbf{IPC}} & 607 & 830 & 26432 & 1284\\
\multirow{1}{*}{\textbf{DMOZ-SMALL}} & 52 & 65 & NS & 168\\
\multirow{1}{*}{\textbf{DMOZ-2010}} & 20190 & 25600 & NS & 42000\\
\multirow{1}{*}{\textbf{DMOZ-2012}} & 50040 & 63000 & NS & 94800\\
\hline
\end{tabular}
\caption{{Total training time (in mins).}} 
\label{table:RuntimeRewire} 
\end{table}

\begin{table}[t!] 
\centering 
\begin{tabular}{|c|c|c|c|} 
\hline
{\bf{$\#$ Executed elementary operation}} & \multicolumn{3}{|c|}{\bf{Dataset}}\\
\cline{2-4}
{\bf{for hierarchy modification}} & \multirow{1}{*}{\textbf{CLEF}} & \multirow{1}{*}{\textbf{DIATOMS}} & \multirow{1}{*}{\textbf{IPC}}\\
\hline
{T-Easy \cite{tang2006acclimatizing}} & \multirow{2}{*}{52} & \multirow{2}{*}{156} & \multirow{2}{*}{412}\\
{(promote, demote, merge)}&&&\\
\hline
{proposed rewHier method} & \multirow{2}{*}{25} & \multirow{2}{*}{34} & \multirow{2}{*}{42}\\
{(NC, PCRewire, ND)}&&&\\
\hline
\end{tabular}
\caption{{Number of elementary operation executed for rewiring approaches.}} 
\label{table:NumHier} 
\end{table}

\begin{figure}[t]
\centering
    \includegraphics[width=0.5\textwidth, height=4.5cm]{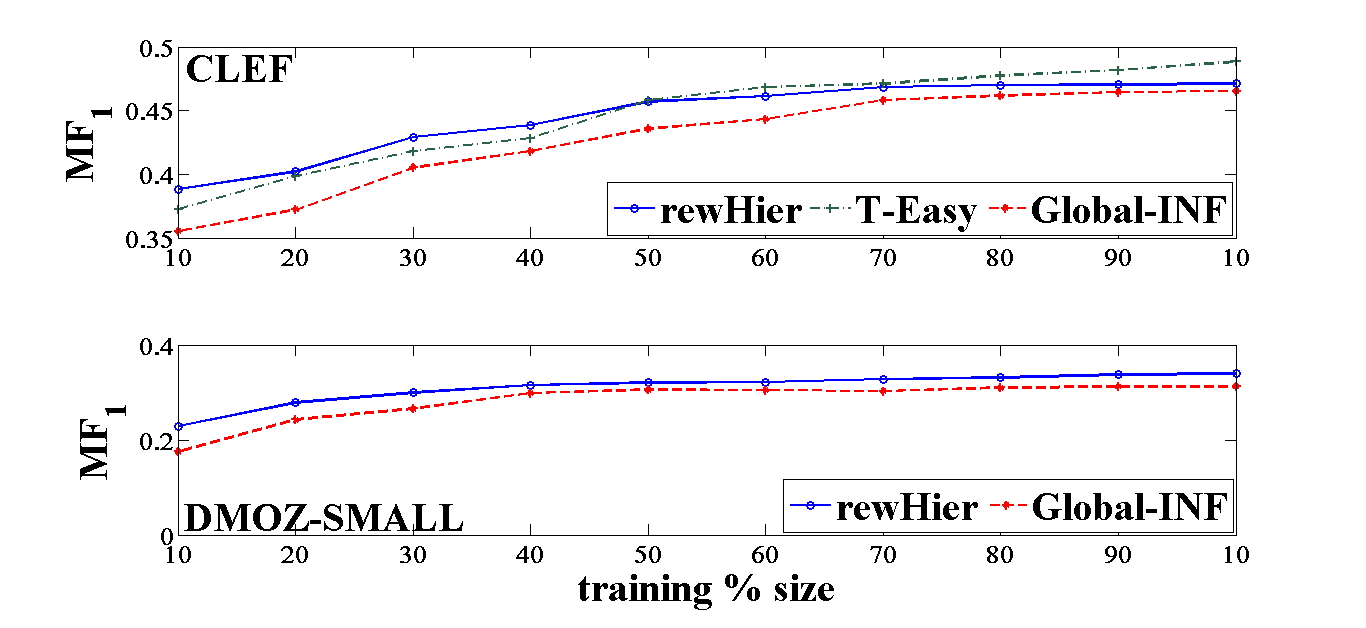}
  \caption{{$MF_1$ performance comparison of rewiring approaches with best method, Global-INF, with varying $\%$ of training size. T-Easy approach is not scalable for DMOZ-SMALL dataset.}}
  \label{RewireFlatten}
\end{figure} 

\subsubsection{Performance based on Hierarchical Metrics}  
Hierarchical evaluation metrics such as 
$hF_1$ computes errors for misclassified 
examples based on
the definition of a defined hierarchy. 
Table \ref{HM} shows the $hF_1$ score for 
the best baseline method, Global-INF and 
the rewiring methods evaluated over the original and the modified 
hierarchy. 
The rewiring methods shows the 
best performance for all the datasets because it is 
able to restructure the hierarchy based on the dataset that is better suited for classification. 

%Moreover, new hierarchy has better performance over original hierarchy because restructuring lead to less misclassifications, resulting in $hF_1$ score improvement. 

\subsubsection{Runtime Comparison}
In Table \ref{table:RuntimeRewire} we compare the training times of the  different models. For training, we learn the models in parallel for 
different classes using multiple compute nodes which are then combined to obtain the final runtime. For our proposed rewiring approach we also 
compute the similarity between different classes in parallel. We can see from  Table \ref{table:RuntimeRewire} that TD-LR takes the the 
least time as there is no  overhead associated with modifying the hierarchy; followed by the Global-INF model which requires retraining  of models
after hierarchy flattening. Rewiring approaches are most expensive because of the 
compute intensive task of either performing similarity computation in 
our proposed approach or multiple hierarchy evaluations using the
T-Easy approach. The 
T-Easy method takes the longest time due to large number of expensive hierarchy evaluations after each elementary operations until the optimal hierarchy is reached. Table \ref{table:NumHier} shows 
the number of elementary operations executed using the T-Easy and the $rewHier$ approach. We can see that T-Easy approach performs 
large number of operations even for smaller datasets (for $e.g.$, 412 operations for IPC datasets in comparison to 42 for the \emph{rewHier}).

\subsection{Effect of varying the Training Size} 
Figure \ref{RewireFlatten} shows the $MF_1$ comparison of rewiring approaches with Global-INF approach on CLEF and DMOZ-SMALL datasets with varying percentage of training size. For both datasets we can see that rewiring approaches outperform
the flattening approaches. For the CLEF dataset with smaller training percentage, 
the \emph{rewHier} approach has better performance. The reason for this behavior might be the over-fitting of the optimal hierarchy with the 
training data in case of T-Easy approach, which results in poor performance on unseen examples. For training dataset with enough examples as expected, T-Easy method gives the best performance but at the cost of expensive runtime. We cannot run T-Easy on the larger DMOZ datasets.

\subsection{Threshold ($\tau$) Selection to Group Similar Classes Pairs}
\label{empDef}
\begin{figure}[t]
\centering
    \includegraphics[width=0.5\textwidth, height=2.5cm]{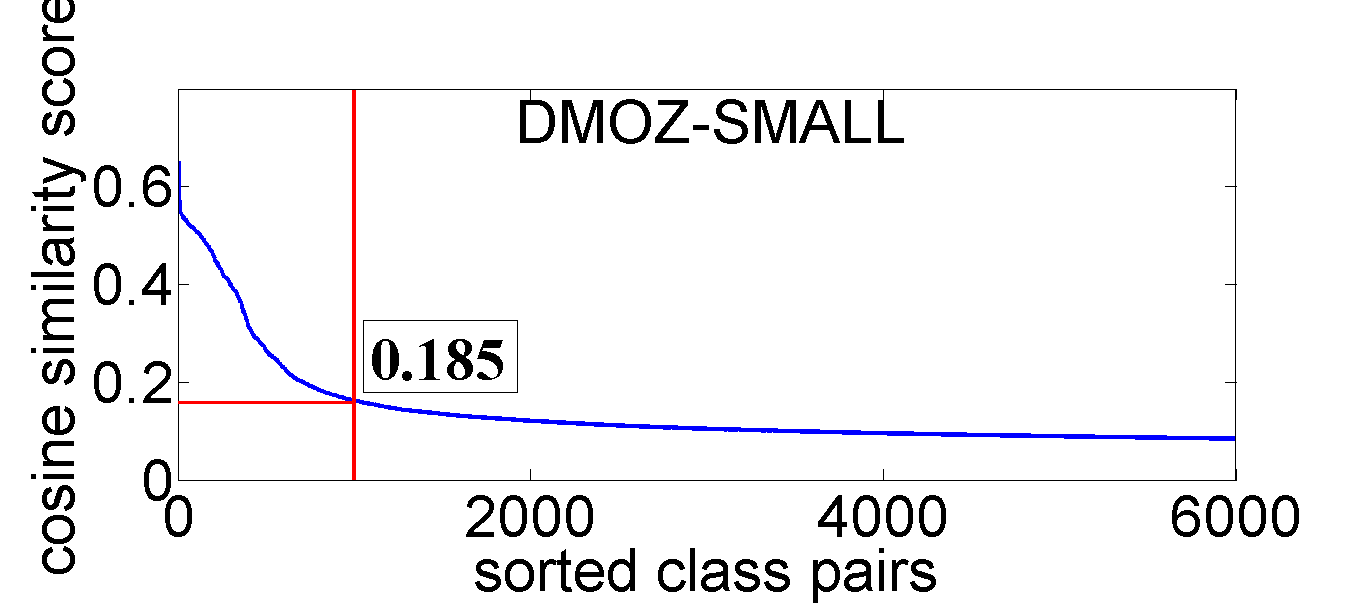}
  \caption{{Sorted cosine similarity scores for DMOZ-SMALL dataset.}}
  \label{SimilarityThreshold}
\end{figure}

%We first sort the cosine similarity scores of class pairs in descending order and than scan to identify the maximum fluctuation point. We set the $\tau$ value as the last scanned score before maximum fluctuation. 
Figure \ref{SimilarityThreshold} shows the sorted (descending order)
class pairs cosine similarity scores for DMOZ-SMALL dataset.
We can see that similarity scores become nearly constant after
1000 pairs (and drops further after 6000, not shown in the
Figure) that does not provide any interesting similar classes
grouping information for taxonomy modification. As such, for this dataset 
choosing threshold  as the similarity score of the
1000-th class pair is a reasonable choice. A similar approach
to determine the threshold is 
applied for other datasets as
well.

\begin{figure}[t!]
    \centering
        \includegraphics[width=0.459\textwidth, height=5.75cm]{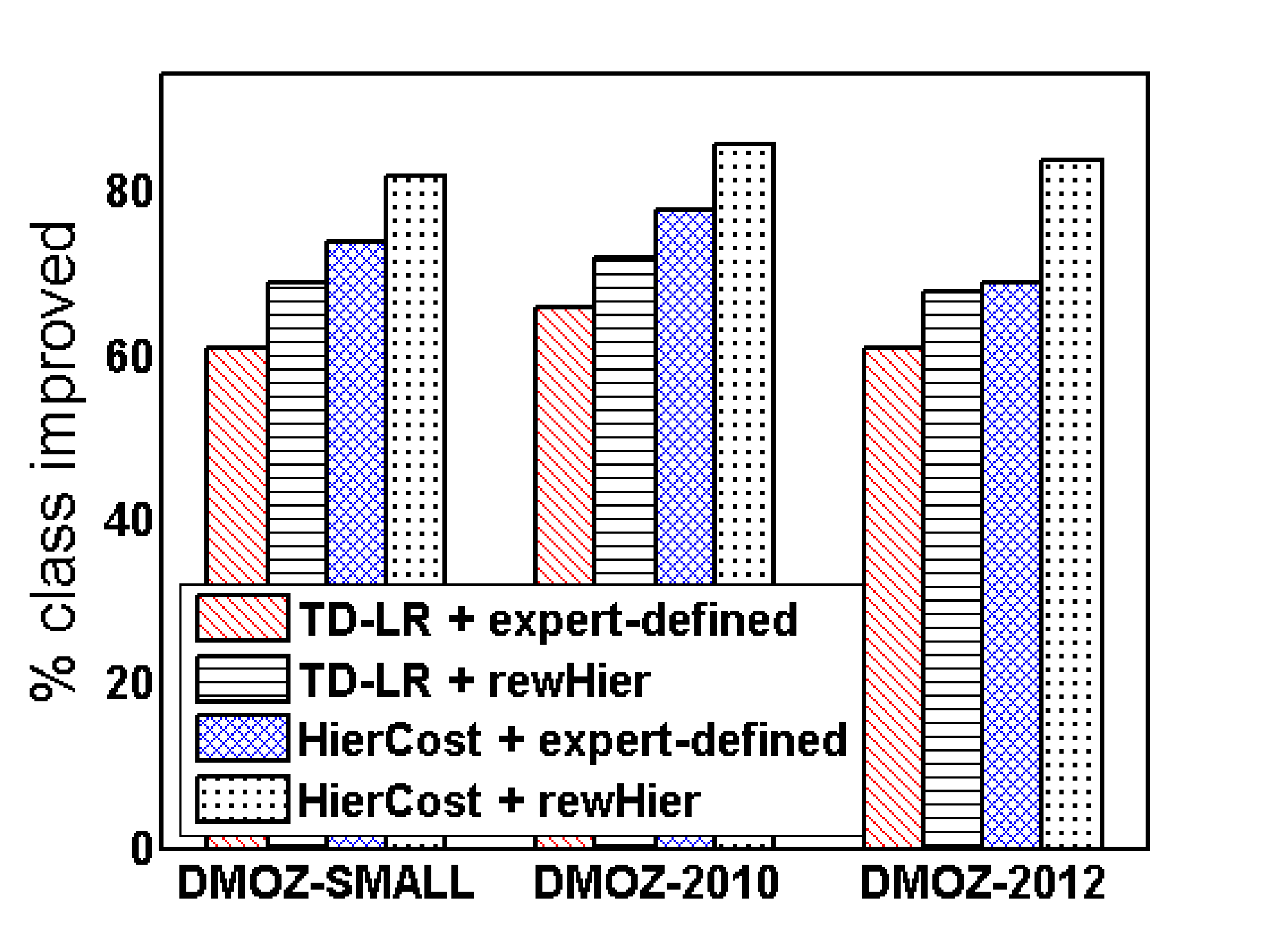}
      \caption{{Percentage of rare categories ($\leq$10 examples per class) classes improved over flat method.}}
  \label{rareImp}
\end{figure}

\begin{table*}[t]
\begin{centering} 
\scriptsize
\begin{tabular}{| l |c c| c c | c c| c c| c c|} 
\hline
\multicolumn{1}{|c|}{\bf{}} &  \multicolumn{2}{c|}{\bf{Flat Method}} & \multicolumn{4}{c|}{\bf{TD-LR}} & \multicolumn{4}{c|}{\bf{HierCost}}\\
\cdashline{4-11}
\multicolumn{1}{|c|}{\bf{Dataset}} &  \multicolumn{2}{c|}{\bf{LR}} & \multicolumn{2}{c|}{\bf{expert-defined}} & \multicolumn{2}{c|}{\bf{rewHier}} & \multicolumn{2}{c|}{\bf{expert-defined}} & \multicolumn{2}{@{\hskip 0.04in}c@{\hskip 0.04in}|}{\bf{rewHier}}\\
\cdashline{2-11}
& \bf{$MF_1$} & \bf{$hF_1$}& \bf{$MF_1$} & \bf{$hF_1$}  & \bf{$MF_1$} & \bf{$hF_1$} & \bf{$MF_1$} & \bf{$hF_1$} & \multicolumn{1}{c}{\bf{$MF_1$}} & \multicolumn{1}{c|}{\bf{$hF_1$}}\\
\hline
\bf{CLEF} &51.31 & 80.58 & 35.92 & 74.52 & 47.10 & 80.14 & 52.30 & 82.18 & {\bf{54.20}} & {\bf{84.42}}\\
\bf{DIATOMS} & 54.17 & 63.50 & 44.46 & 56.15 & 52.14 & 63.24 & 54.16 & 64.13 & {\bf{55.78}} & {\bf{66.31}}\\
\bf{IPC} & 45.74 & 64.00 & 42.51 & 62.57 & 46.04 & 62.57 & 50.10 & 68.45 & {\bf{51.04}} & {\bf{69.43}}\\
\bf{DMOZ-SMALL} & 30.80 & 60.87 & 30.65 & {63.14} & 32.92 & {{66.18}} & 32.98 & 65.58 & {\bf{33.43}} & {\bf{66.30}}\\
\bf{DMOZ-2010} & 27.06 & 53.94 & 28.37 & 54.82 & {29.48} & {56.43} & 29.81 & 58.24 & {\bf{30.35}} & {\bf{58.93}}\\
\bf{DMOZ-2012} & 27.04 & 66.45 & 28.54 & 68.12 & {29.94} & {69.00} & 29.78 & 69.74 & {\bf{30.27}} & {\bf{70.21}}\\
\hline
\end{tabular}
\par\end{centering}
\caption{{Comparative Performance Results.}}
\label{table:hierarchyUseful} 
\end{table*}

\begin{figure*}[t]
    \centering
    \begin{subfigure}[b]{0.459\textwidth}
        \centering
        \includegraphics[width=\textwidth, height=5.75cm]{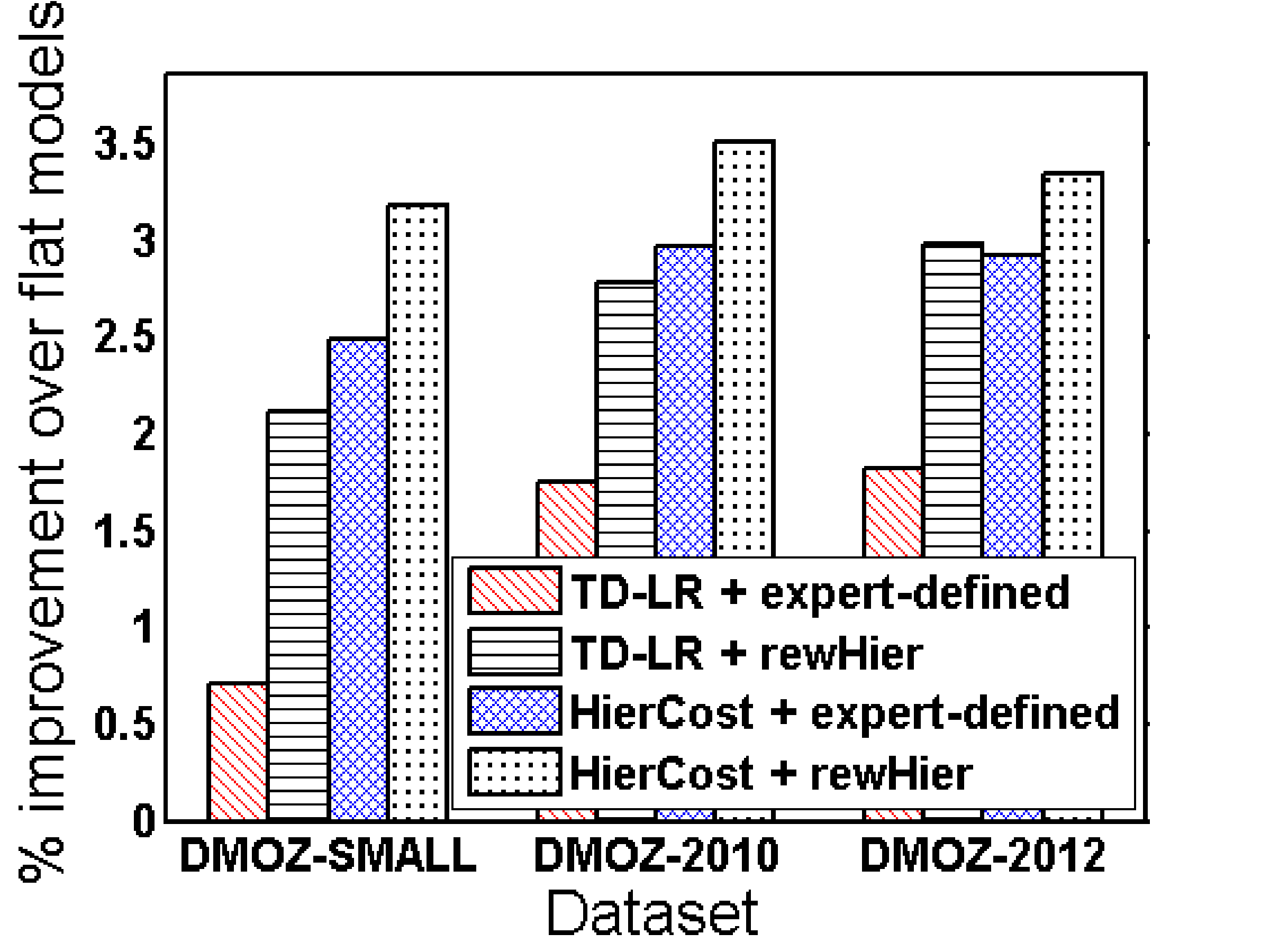}
        \caption{\textbf{$MF_1$}}
    \end{subfigure}
    \hspace*{-5mm}
    \begin{subfigure}[b]{0.459\textwidth}
        \centering
        \includegraphics[width=\textwidth, height=5.75cm]{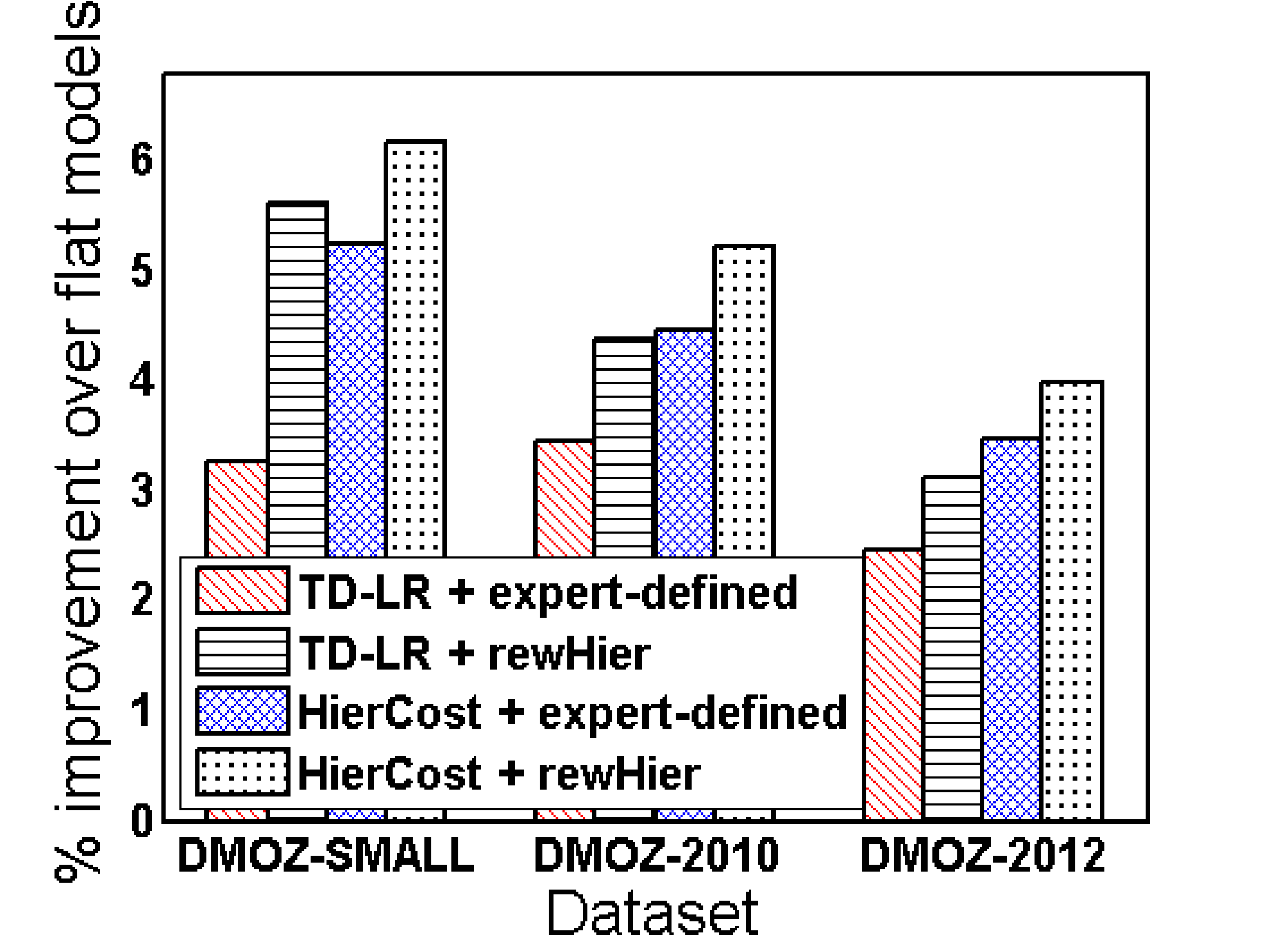}
        \caption{\textbf{$hF_1$}}
    \end{subfigure}
      \caption{{Percentage improvement in $MF_1$ and $hF_1$ scores of various approaches over flat LR approach for classes with rare categories.}}
  \label{rareImp1}
\end{figure*}

\subsection{Improvement over Flat and State-of-the-art Approaches.}
\label{flatCost}
 Figure \ref{rareImp} presents the percentage of classes improved for TD-LR and HierCost HC approaches in comparison to the 
flat approach on DMOZ datasets containing rare categories $i.e.$, less than 10 training examples.  
For the DMOZ-2010 and DMOZ-2012 benchmarks we use a separate held out test dataset since, we do not have the true labels for the provided 
test set used for the online competition. 
From Figure \ref{rareImp} we observe that both the HC approaches outperforms the flat approach irrespective of the hierarchy being used. Rare categories benefit from the utilization of hierarchical relationships, and using the hierarchy improves the accuracy of HC. Moreover, use of \emph{rewHier} to train the TD-LR and HierCost approaches improves 
the classification performance in comparison to using the expert-defined hierarchy. 
Further, the HierCost approach consistently outperforms the TD-LR approach because HierCost penalizes the misclassified instances based on the assignment within the hierarchy. Table \ref{table:hierarchyUseful} gives the more comprehensive results over all classes and Figure \ref{rareImp1} gives $MF_1$ and $hF_1$ improvements for rare categories classes.

In terms of prediction runtime, the TD approaches 
outperform the flat and HierCost approaches. 
The flat and HierCost models invoke all the classifiers 
trained for the leaf nodes to make a prediction 
decision. For the DMOZ-2012 dataset,  the flat and HierCost approaches 
take {$\sim$}220 minutes for 
predicting the labels of test instances, whereas  the
TD-LR model is 3.5 times faster on the same 
hardware configuration.

\section{Related Work}
\label{litReview}
Our work is closely related to the rewiring approach developed in Tang et al. \cite{tang2006acclimatizing}, where the expert-defined hierarchy is 
gradually modified. Iteratively, a subset of the hierarchy is modified and evaluated for classification 
 performance improvement using the HC learning algorithm. Modified changes are retained if the performance results  improve; 
 otherwise the changes are discarded and the process is repeated. This repeated procedure of 
 hierarchy modification continues until the optimal hierarchy is reached. Expensive evaluation at each step makes this approach intractable for large-scale datasets. Another drawback of this approach is deciding which branch of the hierarchy to explore first (for modification) and which 
  elementary operation (promote, demote, merge) to apply at each step. Other work in similar direction can be found in \cite{qi2011hierarchy,nitta2010improving}.

Earlier studies focused on flattening based approaches where some level or nodes are selectively flattened (removed) based on certain criterion \cite{wang2010flatten,babbar2013maximum,malik2010improving}. In other work, learning based approach have been proposed \cite{babbar2013flat}, where nodes to flatten are decided based on classification performance improvement on a validation set. This approach 
although useful for smaller datasets, is not scalable due to the expensive evaluation process after each node removal. Recently, Naik et al. \cite{naik2016inconsistent} proposed a 
taxonomy adaptation where some nodes are intelligently flattened based on empirically defined cut-off threshold and objective function values computed at each node. Hierarchy modification 
using this approach is scalable and beneficial for classification and has been theoretically justified \cite{gao2011discriminative}. 

Other approaches towards hierarchy modification involves generating hierarchy from scratch, 
ignoring the expert-defined hierarchy. These approaches  exploit
hierarchical clustering algorithms for 
generating the hierarchy \cite{li2007hierarchical,punera2005automatically,aggarwal1999merits,chuang2004practical}. Constructing hierarchy 
using clustering approaches is not popular due to its sensitivity to predefined parameters such as number of levels. 

\section{Conclusion and Future Work}
\label{conclusion}
We propose a data-driven filter based rewired approach for hierarchy modification that is more suited for HC. Our method is  robust and can be adapted to work in conjunction with 
any state-of-the-art HC approaches in the literature that utilize hierarchical relationships. Irrespective of the classifiers being trained, our modified hierarchy 
consistently gives better performance over use of clustering or flattening to modify the original  hierarchy. In comparison to previous rewiring approaches, our method gives competitive 
results with much better runtime performance that allow HC approaches to scale to significantly large datasets (e.g., DMOZ). Further, experiments on datasets with skewed 
distribution shows the effectiveness of our proposed method in comparison to flat and state-of-the-art methods, especially for classes with rare categories.  In future,  we plan to study the effect of our method in conjunction with feature selection and other non-linear classification methods. 

\section*{Acknowledgement}
NSF grant \#203337 and \#202882 to Huzefa Rangwala.

\bibliographystyle{unsrt}
\bibliography{TaxMod_reference}

\begin{thebibliography}{10}

\bibitem{cai2004hierarchical}
L.~Cai and T.~Hofmann.
\newblock Hierarchical document categorization with support vector machines.
\newblock In {\em CIKM}, pages 78--87, 2004.

\bibitem{koller1997hierarchically}
D.~Koller and M.~Sahami.
\newblock Hierarchically classifying documents using very few words.
\newblock In {\em ICML}, pages 170--178, 1997.

\bibitem{mccallum1998improving}
A.~McCallum, R.~Rosenfeld, T.~Mitchell, and A.~Ng.
\newblock Improving text classification by shrinkage in a hierarchy of classes.
\newblock In {\em ICML}, pages 359--367, 1998.

\bibitem{dumais2000hierarchical}
S.~Dumais and H.~Chen.
\newblock Hierarchical classification of web content.
\newblock In {\em ACM SIGIR}, pages 256--263, 2000.

\bibitem{sun2001hierarchical}
A.~Sun and E.~Lim.
\newblock Hierarchical text classification and evaluation.
\newblock In {\em ICDM}, pages 521--528, 2001.

\bibitem{liu2005site}
T.~Liu, H.~Wan, T.~Qin, Z.~Chen, Y.~Ren, and W.~Ma.
\newblock Site abstraction for rare category classification in large-scale web
  directory.
\newblock In {\em WWW: Special interest tracks \& posters}, pages 1108--1109,
  2005.

\bibitem{naik2016inconsistent}
A.~Naik and H.~Rangwala.
\newblock Inconsistent node flattening for improving top-down hierarchical
  classification.
\newblock In {\em IEEE DSAA}, 2016.

\bibitem{li2007hierarchical}
T.~Li, S.~Zhu, and M.~Ogihara.
\newblock Hierarchical document classification using automatically generated
  hierarchy.
\newblock {\em JIIS}, 29(2):211--230, 2007.

\bibitem{xiao2011hierarchical}
L.~Xiao, D.~Zhou, and M.~Wu.
\newblock Hierarchical classification via orthogonal transfer.
\newblock In {\em ICML}, pages 801--808, 2011.

\bibitem{zimek2010study}
A.~Zimek, F.~Buchwald, E.~Frank, and S.~Kramer.
\newblock A study of hierarchical and flat classification of proteins.
\newblock {\em IEEE/ACM TCBB}, 7(3):563--571, 2010.

\bibitem{tang2006acclimatizing}
L.~Tang, J.~Zhang, and H.~Liu.
\newblock Acclimatizing taxonomic semantics for hierarchical content
  classification.
\newblock In {\em ACM SIGKDD}, pages 384--393, 2006.

\bibitem{qi2011hierarchy}
X.~Qi and B.~Davison.
\newblock Hierarchy evolution for improved classification.
\newblock In {\em CIKM}, pages 2193--2196, 2011.

\bibitem{charuvaka2015}
A.~Charuvaka and H.~Rangwala.
\newblock Hiercost: Improving large scale hierarchical classification with cost
  sensitive learning.
\newblock In {\em ECML PKDD}, 2015.

\bibitem{babbar2013maximum}
R.~Babbar, I.~Partalas, E.~Gaussier, and MR. Amini.
\newblock Maximum-margin framework for training data synchronization in
  large-scale hierarchical classification.
\newblock In {\em Neural Information Processing}, pages 336--343, 2013.

\bibitem{wang2010flatten}
X.~Wang and B.~Lu.
\newblock Flatten hierarchies for large-scale hierarchical text categorization.
\newblock In {\em ICDIM}, pages 139--144, 2010.

\bibitem{babbar2013flat}
R.~Babbar, I.~Partalas, E.~Gaussier, and M.~Amini.
\newblock On flat versus hierarchical classification in large-scale taxonomies.
\newblock In {\em NIPS}, pages 1824--1832, 2013.

\bibitem{punera2005automatically}
K.~Punera, S.~Rajan, and J.~Ghosh.
\newblock Automatically learning document taxonomies for hierarchical
  classification.
\newblock In {\em WWW: Special interest tracks \& posters}, 2005.

\bibitem{malik2010improving}
H~Malik.
\newblock Improving hierarchical svms by hierarchy flattening and lazy
  classification.
\newblock In {\em Large-Scale HC Workshop of ECIR}, 2010.

\bibitem{nitta2010improving}
K.~Nitta.
\newblock Improving taxonomies for large-scale hierarchical classifiers of web
  docs.
\newblock In {\em CIKM}, pages 1649--1652, 2010.

\bibitem{steinbach2004challenges}
M.~Steinbach, L.~Ert{\"o}z, and V.~Kumar.
\newblock The challenges of clustering high dimensional data.
\newblock In {\em New Directions in Statistical Physics}, pages 273--309. 2004.

\bibitem{gopal2013recursive}
S.~Gopal and Y.~Yang.
\newblock Recursive regularization for large-scale classification with
  hierarchical \& graphical dependencies.
\newblock In {\em ACM SIGKDD}, pages 257--265, 2013.

\bibitem{dimitrovski2011hierarchical}
I.~Dimitrovski, D.~Kocev, S.~Loskovska, and S.~D{\v{z}}eroski.
\newblock Hierarchical annotation of medical images.
\newblock {\em Pattern Recognition}, 44(10):2436--2449, 2011.

\bibitem{Dimitrovski11:jrnl}
I.~Dimitrovski, D.~Kocev, S.~Loskovska, and S.~D\v{z}eroski.
\newblock Hierarchical classification of diatom images using predictive
  clustering trees.
\newblock {\em Ecological Informatics}, 7:19--29, 2012.

\bibitem{yang1999re}
Y.~Yang and X.~Liu.
\newblock A re-examination of text categorization methods.
\newblock In {\em ACM SIGIR}, pages 42--49, 1999.

\bibitem{gao2011discriminative}
T.~Gao and D.~Koller.
\newblock Discriminative learning of relaxed hierarchy for large-scale visual
  recognition.
\newblock In {\em ICCV}, pages 2072--2079, 2011.

\bibitem{aggarwal1999merits}
C.~Aggarwal, S.~Gates, and P.~Yu.
\newblock On the merits of building categorization systems by supervised
  clustering.
\newblock In {\em SIGKDD}, pages 352--356, 1999.

\bibitem{chuang2004practical}
S.~Chuang and L.~Chien.
\newblock A practical web-based approach to generating topic hierarchy for text
  segments.
\newblock In {\em CIKM}, pages 127--136, 2004.

\end{thebibliography}

\end{document}